\pdfoutput=1

\documentclass[11pt]{article}

\usepackage[preprint]{acl}

\usepackage{times}
\usepackage{latexsym}
\usepackage{url}
\usepackage{booktabs}
\usepackage{adjustbox}
\usepackage{colortbl}
\usepackage{threeparttable}
\usepackage{makecell}
\usepackage{multirow} 
\usepackage{subcaption}
\usepackage{amsmath}
\usepackage{amssymb}
\usepackage{amsthm}
\usepackage{capt-of}
\usepackage{soul}
\usepackage{pifont}

\usepackage[T1]{fontenc}

\usepackage[utf8]{inputenc}

\usepackage{microtype}

\usepackage{inconsolata}

\usepackage{graphicx}

\title{MoDification: Mixture of Depths Made Easy}

\author{Chen Zhang\textsuperscript{\ding{168}}, Meizhi Zhong\textsuperscript{\ding{169}}, Qimeng Wang\textsuperscript{\ding{170}}, Xuantao Lu\textsuperscript{\ding{170}}, Zheyu Ye\textsuperscript{\ding{170}}, \\
  \textbf{Chengqiang Lu\textsuperscript{\ding{170}}, Yan Gao\textsuperscript{\ding{170}}, Yao Hu\textsuperscript{\ding{170}}, Kehai Chen\textsuperscript{\ding{169}}, Min Zhang\textsuperscript{\ding{169}}, Dawei Song\textsuperscript{\ding{168}}\Thanks{ Dawei Song is the corresponding author.}} \\
  \textsuperscript{\ding{168}}Beijing Institute of Technology \\
  \textsuperscript{\ding{169}}Harbin Institute of Technology, Shenzhen \\
  \textsuperscript{\ding{170}}Xiaohongshu \\
  \texttt{chenzhang9702@outlook.com} \\}

\begin{document}
\maketitle
\begin{abstract}
Long-context efficiency has recently become a trending topic in serving large language models (LLMs). And mixture of depths (MoD) is proposed as a perfect fit to bring down both latency and memory. In this paper, however, we discover that MoD can barely transform existing LLMs without costly training over an extensive number of tokens. To enable the transformations from any LLMs to MoD ones, we showcase \textsf{top-k} operator in MoD should be promoted to \textsf{threshold-p} operator, and refinement to architecture and data should also be crafted along. All these designs form our method termed MoDification. Through a comprehensive set of experiments covering model scales from 3B to 70B, we exhibit MoDification strikes an excellent balance between efficiency and effectiveness. MoDification can achieve up to $\sim$1.2$\times$ speedup in latency and $\sim$1.8$\times$ reduction in memory compared to original LLMs especially in long-context applications.
\end{abstract}

\section{Introduction}

Long-context efficiency is turning to be one of the core concerns in large language model (LLM) serving~\citep{DBLP:journals/corr/TouvronLI23,DBLP:journals/corr/TouvronMS23,DBLP:journals/corr/DubeyJP24,DBLP:journals/corr/BaiBC23,DBLP:journals/corr/YangYH24}. Typically, a long context can incur dramatically huge latency and memory overhead either at prefilling or decoding stage~\citep{DBLP:journals/tmlr/Wan0LA0LQYZZC024}.

To address this consumption, strategies like speculative decoding~\citep{DBLP:conf/icml/LeviathanKM23,DBLP:journals/corr/ChenBI23,DBLP:conf/coling/LiuZ024} and key-value (KV) cache compression~\citep{DBLP:conf/iclr/XiaoTCHL24,DBLP:journals/corr/LiHY24,XiaoTZ24} have already been utilized to enhance the latency and memory, respectively. However, dedicating efforts to separate components is not always ideal. As a consequence, mixture of depths~\citep[MoD,][]{DBLP:journals/corr/RaposeRR24} has recently been introduced to simultaneously consider both two facets. Conceptually, the key feature of MoD is enabling LLMs to conditionally eliminate computations of layers over certain tokens. And distinguished from early exiting~\citep[EE,][]{DBLP:conf/icml/ChenPLDZ24}, MoD skips over rather than exiting from intermediate layers so that performance is better preserved.

Frustratingly, MoD can hardly guarantee both efficiency and effectiveness without extremely costly training. To be more specific, current advances of MoD are only observed in cases where training is started from scratch, predominately narrowing the impact of MoD.

\begin{figure}
    \centering
    \includegraphics[width=0.47\textwidth]{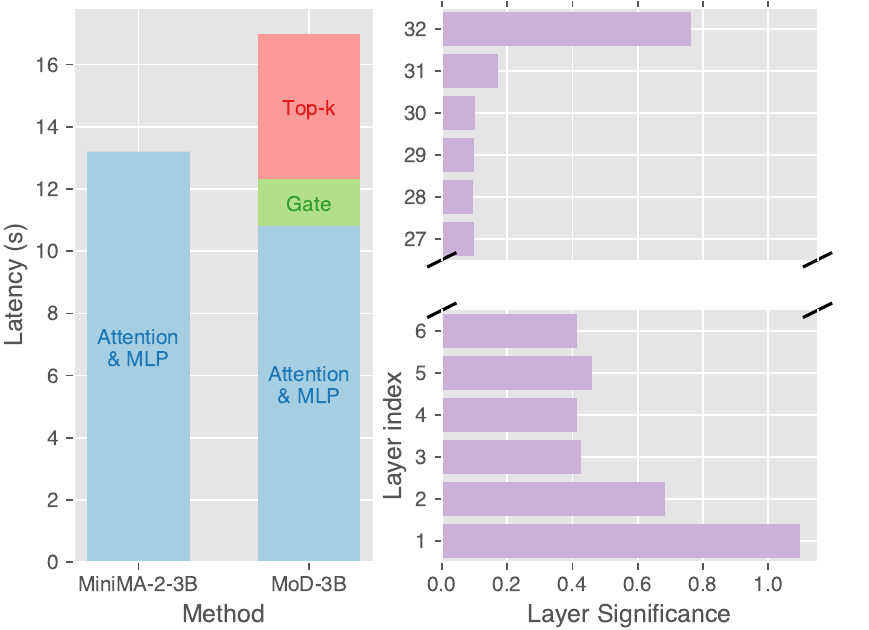}
    \caption{The sub-optimality of MoD due to the use of \textsf{top-k} operator. For efficiency, the efficiency improvement is restricted because \textsf{top-k} operator is not cheap and unimportant computations are preserved by retaining a constant number of tokens at every layer. For effectiveness, the effectiveness is undesired because dominated computations are saved by skipping a constant number of tokens at every layer.}
    \label{fig:mod_suboptimality}
\end{figure}

The sub-optimality is mainly due to the improper use of \textsf{top-k} operator. The \textsf{top-k} operator is basically employed at each layer to determine which tokens should be skipped in the computation of the layer. Unfortunately, for efficiency, the \textsf{top-k} operator is computationally expensive~\citep{DBLP:conf/hpca/WangZH21} and forces a constant number of tokens retained, possibly restricting the efficiency improvement brought by computation saving at one unimportant layer, as shown in Figure~\ref{fig:mod_suboptimality} (left). For effectiveness, it deviates from the distribution of layer significance~\citep{DBLP:journals/corr/MenXZ24}, potentially resulting in undesired computation saving at one dominated layer, as shown in Figure~\ref{fig:mod_suboptimality} (right). These disadvantages largely impede the applications of MoD.

In this paper, in order to broaden the scope of MoD, we target at converting existing LLM checkpoints to MoD ones with our designed MoDification.\footnote{MoDification is a re-invented compound representing the process of adapting a LLM to a MoD one, following the naming convention of MoEfication~\citep{DBLP:conf/acl/ZhangL00S022}.} The designs of MoDification follow two guidelines: 1) alleviating the sub-optimality of MoD, and 2) minimizing the training of MoD. Under the guidelines, we firstly use \textsf{threshold-p} operator as an drop-in replacement of the \textsf{top-k} operator. On one hand, \textsf{threshold-p} operator is not much computationally expensive and allows any number of retained tokens; on the other hand, it can make the layer itself determine the priority of tokens through adaptive gates across layers. We secondly find around 10B tokens with decent diversity are fairly enough in the conversion process.

We conduct experiments on model scales ranging from 3B to 70B. The efficiency and effectiveness of MoDification are verified on such a wide array of scales. With acceptable performance decline, MoDification can achieve up to $\sim$1.2$\times$ speedup in latency and $\sim$1.8$\times$ reduction in memory compared to original LLMs especially in long-context applications. On the contrary, MoD can unexpectedly lead to slow-down in latency under the same experimental settings. Further ablations and discussions validate the trustworthiness of the design choices in MoDification.

\section{Background}

\paragraph{Long-context Efficiency}

Over the years, the super expressiveness of large language models (LLMs) has been widely witnessed thanks to the stable transformer architecture~\citep{DBLP:conf/nips/VaswaniSPUJGKP17} and the massive pretraining~\citep{DBLP:conf/nips/BrownMRSKDNSSAA20}. Concretely, two modules are involved in a transformer layer, namely, self-attention (Attention) module and multi-layer perceptron (MLP) module. Essentially, layer norm (Norm) and residual connection are also armed around. Given the hidden state of the $i$-th token $\mathbf{x}_i$, the forward process in one transformer layer can usually be depicted as below:
\begin{equation}
\begin{aligned}
    \mathbf{y}_i=\text{Attention}(\text{Norm}(\mathbf{x}_i))+\mathbf{x}_i, \\
    \mathbf{z}_i=\text{MLP}(\text{Norm}(\mathbf{y}_i))+\mathbf{y}_i, 
\end{aligned}
\end{equation}

While the transformer architecture admits the scaling of pretraining, a critical bottleneck during inference is also revealed by the transformer. In other words, the transformer architecture mainly grants parallelism during training, yet only generates tokens one followed by another and requires KV caches adequately stored during inference. This bottleneck may easily slow down the expected speed and shoot over the provided hardware, especially for long-context circumstances~\citep{DBLP:conf/iclr/XiaoTCHL24}.

To this demand, many approaches have been proposed to either counter the latency or the memory increment ~\citep{DBLP:journals/tmlr/Wan0LA0LQYZZC024}. Among these approaches, speculative decoding~\citep{DBLP:journals/corr/ChenBI23} is a representative cluster in reducing the latency and KV cache compression~\citep{DBLP:conf/iclr/XiaoTCHL24} is another symbolic group in mitigating the memory. However, for an integrated inference system, it is might be somehow redundant to conquer the latency and the memory separately. Consequently, later solutions to some extent shift the focus to conditional computation~\citep{DBLP:journals/corr/BengioBPP15}, in which early exiting~\citep[EE,][]{DBLP:conf/icml/ChenPLDZ24} and mixture of depths~\citep[MoD,][]{DBLP:journals/corr/RaposeRR24} are of central roles. 

\paragraph{Mixture of Depths}

In a nutshell, MoD attempts to eliminate unnecessary layer computations over tokens via a token-level gate (Gate) together with a \textsf{top-k} operator (Top-k). In this way, the forward process within one transformer layer is changed as below:
\begin{equation}
\label{eq:mod}
\begin{aligned}
    g_i=\text{Gate}(\mathbf{x}_i),\quad f_i=\text{Top-k}(g_i, k), \\
    \mathbf{y}_i=
    \begin{cases}
        \text{Attention}(\text{Norm}(\mathbf{x}_i))+\mathbf{x}_i, & f_i=1, \\
        \mathbf{x}_i, & f_i=0, \\
    \end{cases} \\
    \mathbf{z}_i=
    \begin{cases}
        g_i\cdot\text{MLP}(\text{Norm}(\mathbf{y}_i))+\mathbf{y}_i, & f_i=1, \\
        \mathbf{y}_i, & f_i=0, \\
    \end{cases} \\
\end{aligned}
\end{equation}
where $g_i\in[0,1]$ is a normalized score and $f_i$ is an indicator of whether the $i$-th token is exactly ranked among the Top-k tokens regarding its score yielded by the Gate. In addition, MoD is usually applied in an interleaved manner, that is, only one out of two adjacent layers is equipped with MoD.

\begin{figure*}
    \centering
    \includegraphics[width=0.9\textwidth]{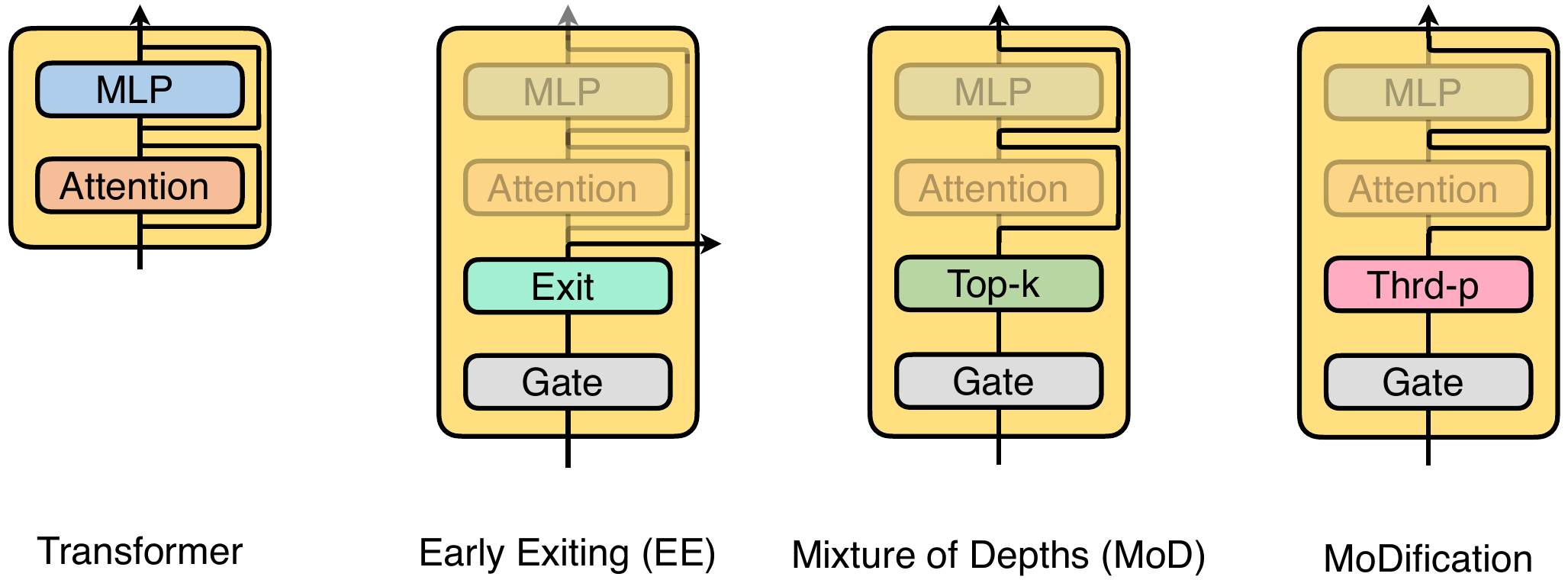}
    \caption{The comparison among transformer, early exiting (EE), mixture of depths (MoD), and our MoDification.}
    \label{fig:comparison}
\end{figure*}

It is acknowledged that EE is a fine alternative of MoD. Instead of skipping over intermediate layers, EE aggressively exits from intermediate layers and directly ignores later layers, as in Figure~\ref{fig:comparison}. Because of this, it is claimed that MoD is more flexible than EE does, and can offer better performance as well.

Despite the superiority of MoD, it does not come without pain. Although up-to-date progress~\citep{DBLP:journals/corr/RaposeRR24} has indeed supported that MoD delivers great success, the training is always launched from scratch with an enormous amount of data. We argue training MoD with an appropriate quantity of data and from publicly released checkpoints should likewise be a pivotal track that is worthy of exploration. And through easing the training, MoD might fully shed its light on related areas.

In our pilot study, we have uncovered sub-optimality pertained to both efficiency and effectiveness of MoD in above lite training setting. Primarily, the sub-optimality is concerned with the improper use of the \textsf{top-k} operator. From the efficiency perspective, the \textsf{top-k} operator is computationally expensive~\citep{DBLP:conf/hpca/WangZH21} and forces a constant number of tokens retained, and can limit computation saving thus efficiency improvement for unimportant layers. From the effectiveness perspective, the \textsf{top-k} operator suffers a divergence from the true distribution of layer significance~\citep{DBLP:journals/corr/MenXZ24}, and can undesirably result in excessive computation saving for dominated layers. The evidence can be referred to in Figure~\ref{fig:mod_suboptimality}.

\section{MoDification}

In this work, we put forward such a method, named MoDification, that is aimed at converting existing LLM checkpoints to MoD ones with slight training compute. 

To circumvent the sub-optimality of original MoD, two designs are engaged, in which one is associated with architecture and the other is connected with data. The first one is to use \textsf{threshold-p} operator in place of the \textsf{top-k} operator, as shown in Figure~\ref{fig:comparison}. \textsf{threshold-p} operator is not much computationally expensive and allows any number of retained tokens. We then instantly discover this \textsf{threshold-p} replacement correlates MoDification to mixture of experts (MoE)~\citep{DBLP:conf/iclr/ShazeerMMDLHD17,DBLP:journals/jmlr/FedusZS22} and we can empower MoDification with cutting-edge techniques that have been widely used. The second one is to enhance the data diversity and constrain the training to the scale of 10B tokens. We luckily unveil this data scale is fairly enough in practice.

\paragraph{Architecture}

After applying \textsf{threshold-p} operator to Equation~\ref{eq:mod}, we make a further refinement to the formulation. The formula can now be briefly formed as below:
\begin{equation}
\begin{aligned}
    g_i=\text{Gate}(\mathbf{x}_i),\quad f_i=\text{Threshold-p}(g_i, p), \\
    \mathbf{y}_i=
    \begin{cases}
        g_i\cdot\text{Attention}(\text{Norm}(\mathbf{x}_i))+\mathbf{x}_i, & f_i=1, \\
        \mathbf{x}_i, & f_i=0, \\
    \end{cases} \\
    \mathbf{z}_i=
    \begin{cases}
        g_i\cdot\text{MLP}(\text{Norm}(\mathbf{y}_i))+\mathbf{y}_i, & f_i=1, \\
        \mathbf{y}_i, & f_i=0, \\
    \end{cases} \\
\end{aligned}
\end{equation}
where the refinement is imposed as also multiplying the gate value $g_i$ to the Attention module. Plus to this, MoDification also follows the interleaved fashion as MoD does.

In fact, the conditional provided by $f_i$ can be interpreted from an MoE view as below:
\begin{equation}
\mathbf{z}_i=
\begin{cases}
    g_i\cdot\text{MLP}(\text{Norm}(\mathbf{y}_i))+\mathbf{y}_i, & f_i=1, \\
    (1-g_i)\cdot\text{NoOp}(\mathbf{y}_i)+\mathbf{y}_i, & f_i=0, \\
\end{cases}
\end{equation}
where NoOp generally means no operation is carried out and always leads to zero output.

Thereby, MoDification is fundamentally a two-expert MoE where one of the two experts is NoOp. In this sense, the $p$ value of the \textsf{threshold-p} operator should theoretically be 0.5, which is henceforth leveraged unless otherwise specified. Moreover, to promote the sparsity of MoDification, we are inspired by the expert load balancing technique~\citep{DBLP:journals/jmlr/FedusZS22} in MoE and suggest a layer load reducing objective as below:
\begin{equation}
\begin{aligned}
    F_{j}=\mathbb{E}_{\mathbf{x}_{i}}[f_i],\quad G_{j}=\mathbb{E}_{\mathbf{x}_{i}}[g_i], \\
    \mathcal{R}=\alpha\cdot\sum^{L}_{j=1}F_{j}\cdot G_{j},
\end{aligned}
\end{equation}
where $\alpha$ is a coefficient that should be manually tuned, and subscript $j$ denotes the $j$-th layer among overall $L$ layers. While $F_{j}$ depicts the fraction of tokens dispatched to the layer, $G_{j}$ describes the fraction of the routing probability to the layer. And reducing $R$ principally triggers more tokens be assigned to the NoOp and thus skipped by the layer. This reducing objective, together with the language modeling objective $\mathcal{L}$ that drives necessarily layer load preserving, makes MoDification a reasonable game where LLMs learn to strike a good balance between efficiency and effectiveness.

\paragraph{Data}

\begin{table}[ht]
    \centering
    \caption{The data mixture used to conduct MoDification.}
    \begin{adjustbox}{width=0.41\textwidth,center}
    \begin{tabular}{lrr}
    \toprule
        \textbf{Data} & \textbf{Tokens} & \textbf{Proportion} \\
    \midrule
        CCI~\citeyearpar{BAAI23} & 1.5B & 14.3\% \\
        Wikipedia~\citeyearpar{DBLP:conf/naacl/DevlinCLT19} & 0.2B & 1.9\% \\
        BookCorpus~\citeyearpar{DBLP:conf/iccv/ZhuKZSUTF15} & 0.1B & 1.0\% \\
        Cosmopedia~\citeyearpar{BenAllalLA24} & 1.0B & 9.5\% \\
        FineWeb-Edu~\citeyearpar{Lozhkov24} & 4.5B & 42.9\% \\
        RedPajama-Stack~\citeyearpar{Together23} & 0.5B & 4.8\% \\
        RedPajama-GitHub~\citeyearpar{Together23} & 1.3B & 12.3\% \\
        Proof-Pile~\citeyearpar{Azerbayev23} & 1.3B & 12.3\% \\
        FLAN~\citeyearpar{DBLP:journals/jmlr/ChungHLZTFL00BW24} & 0.1B & 1.0\% \\
    \midrule
        Total & 10.5B & 100.0\% \\
    \bottomrule
    \end{tabular}
    \end{adjustbox}
    \label{tab:data}
\end{table}

With the aim of reducing training time, we try to incorporate diverse domains of data to compensate limited numbers of data points. The data mixture can be found in Table~\ref{tab:data}.

The proposed data mixture generally covers domains like webs, wikis, books, codes, maths, etc. In doing so, we expect MoDification could be trained to approximate original mixtures of any existing LLM checkpoints.

\section{Experiments}

\paragraph{Setup}

We benchmark MoDification mainly against MoD, and also compare these two to original LLM. We conduct evaluation upon MiniMA-2-3B~\citep{DBLP:journals/corr/ZhangSY23} and LLaMA-2-\{7,13,70\}B~\citep{DBLP:journals/corr/TouvronMS23} to examine the scalability of MoDification. For efficiency measures, we consider average latency in second and memory in gigabyte across a wide array of prefilling and decoding lengths ranging from 64 to 2,048 as key metrics. For effectiveness measures, we test above methods on commonly cared datasets including MMLU~\citep{DBLP:conf/iclr/HendrycksBBZMSS21}, CEval~\citep{DBLP:conf/nips/HuangBZZZSLLZLF23}, DROP~\citep{DBLP:conf/naacl/DuaWDSS019}, BBH~\citep{DBLP:conf/acl/SuzgunSSGTCCLCZ23}, HumanEval~\citep{DBLP:journals/corr/ChenTJ21}, and GSM8k~\citep{DBLP:journals/corr/CobbeKB21}. The core metrics reported on these datasets are either accuracy, exact matching score, or pass@1 according to the designs.

\paragraph{Implementation}

We train MoDification and MoD with an overall batch size of 1,024 and a sequence length of 4,096. This leads to 4M tokens per optimization step. The learning rate is 3e-5, and weight decay is 1e-1. The learning rate is scheduled with a linear warm-up and a cosine decay, where the warm-up stage takes 1\% optimization steps and the decay stage takes all the left optimization steps. The gradients will be properly clipped to keep the norm under 5e-1. The training precision is bfloat16. DeepSpeed~\citep{DBLP:conf/kdd/RasleyRRH20}, FlashAttention~\citep{DBLP:conf/nips/DaoFERR22}, and gradient checkpointing~\citep{DBLP:conf/icml/ChenPLDZ24} are necessarily enabled to make the training possible with 16 Nvidia A100 GPUs.

The $k$ for the \textsf{top-k} operator is 512 by default, and the $p$ for the \textsf{threshold-p} operator is 0.5 for MiniMA-2-3B as mentioned earlier but 0.55 for LLaMA-2-\{7,13,70\}B since slightly larger threshold is demanded for larger LLMs in our trials. The best coefficient $\alpha$ for layer load reducing objective is 0.01 in our developments. 

\begin{table*}[ht]
    \centering
    \caption{The main comparison results concerning both efficiency and effectiveness. Better results among the ones reported from MoD and MoDification are \textbf{boldfaced} at varying blocks. Results enabled with INT8 quantization~\citep{DBLP:journals/corr/DettmersLB22} are marked with \textsuperscript{\dag}.}
    \begin{adjustbox}{width=0.99\textwidth,center}
    \begin{tabular}{lcccccccc}
    \toprule
        \textbf{Method} & \makecell[c]{\textbf{Latency}\\\textcolor{gray}{Second}} & \makecell[c]{\textbf{Memory}\\\textcolor{gray}{GiB}} & \makecell[c]{\textbf{MMLU}\\\textcolor{gray}{5-shot Acc}} & \makecell[c]{\textbf{CEval}\\\textcolor{gray}{5-shot Acc}} & \makecell[c]{\textbf{DROP}\\\textcolor{gray}{3-shot EM}} & \makecell[c]{\textbf{BBH}\\\textcolor{gray}{3-shot EM}} & \makecell[c]{\textbf{HumanEval}\\\textcolor{gray}{0-shot Pass@1}} & \makecell[c]{\textbf{GSM8k}\\\textcolor{gray}{8-shot CoT EM}} \\
    \midrule
        MiniMA-2-3B & 11.0 & 7.4 & 40.1 & 44.7 & 23.1 & 31.4 & 8.9 & 14.6 \\
        \quad - MoD & 16.2 & \textbf{6.7} & 25.8 & 25.7 & 14.0 & 27.7 & 0.0 & 2.2 \\
        \quad - MoDification & \textbf{10.9} & \textbf{6.7} & \textbf{30.0} & \textbf{32.2} & \textbf{20.5} & \textbf{30.8} & \textbf{9.8} & \textbf{5.7} \\
    \midrule
        LLaMA-2-7B & 14.5 & 16.2 & 45.0 & 33.1 & 33.1 & 32.1 & 13.4 & 12.1  \\
        \quad - MoD & 21.6 & 14.9 & 25.5 & 26.0 & 1.8 & 7.0 & 0.0 & 0.0 \\
        \quad - MoDification & \textbf{13.5} & \textbf{14.7} & \textbf{36.4} & \textbf{28.6} & \textbf{25.8} & \textbf{30.0} & \textbf{8.9} & \textbf{10.3} \\
    \midrule
        LLaMA-2-13B & 19.4 & 30.3 & 54.4 & 41.2 & 43.4 & 38.1 & 14.0 & 24.1 \\
        \quad - MoD & 30.9 & 28.2 & 25.7 & 24.6 & 1.5 & 7.8 & 0.0 & 0.0 \\
        \quad - MoDification & \textbf{18.1} & \textbf{27.9} & \textbf{44.4} & \textbf{37.3} & \textbf{34.2} & \textbf{34.1} & \textbf{11.0} & \textbf{21.2} \\
    \midrule
        LLaMA-2-70B & 201\textsuperscript{\dag} & 72\textsuperscript{\dag} & 68.6 & 53.9 & 63.3 & 51.6 & 28.7 & 53.4 \\
        \quad - MoD & 276\textsuperscript{\dag} & 71\textsuperscript{\dag} & 41.2 & 30.1 & 2.5 & 12.2 & 0.0 & 0.0 \\
        \quad - MoDification & 178\textsuperscript{\dag} & 71\textsuperscript{\dag} & \textbf{54.6} & \textbf{46.9} & \textbf{43.6} & \textbf{38.3} & \textbf{15.2} & \textbf{38.0} \\
    \bottomrule
    \end{tabular}
    \end{adjustbox}
    \label{tab:comparison}
\end{table*}

\begin{figure}[t]
    \centering
    \includegraphics[width=0.47\textwidth]{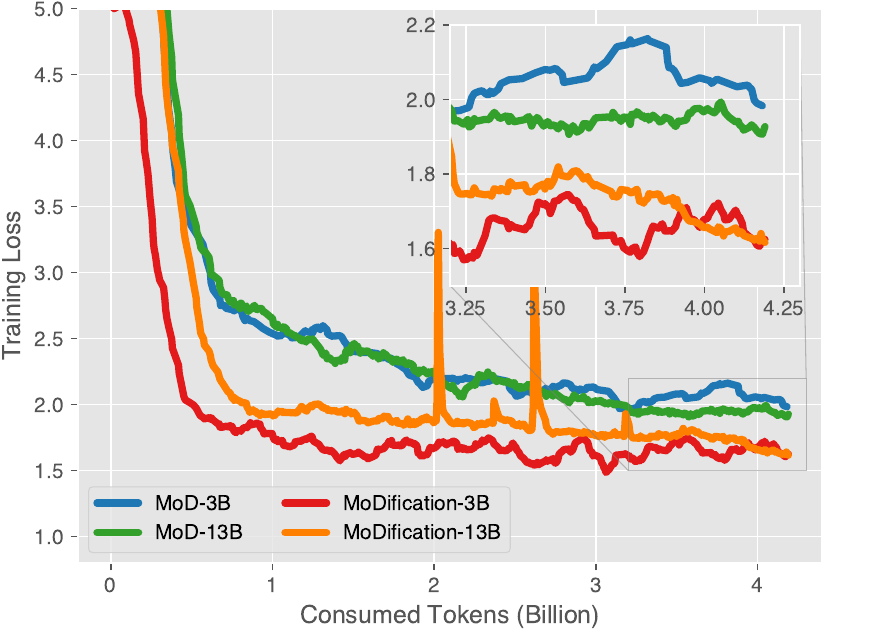}
    \caption{The training losses. The steady loss curves indicate that both MoD and MoDification are rigorously optimized.}
    \label{fig:loss}
\end{figure}

\paragraph{Results}

From the main comparison results between MoD and MoDification in Table~\ref{tab:comparison}, we unearth that MoDification is a way better choice than MoD does. For efficiency, MoDification can reduce both the latency and memory consumption while MoD can unexpectedly increase the latency. For effectiveness, MoDification also enjoys better results and and preserves most performance from original considered LLMs. The comparison showcases the superiority of MoDification over MoD. And the realized efficiency merits seem to be positively correlated to the model scales in practice.

One may think that MoD is not properly configured in training and thus shows very unstable and almost meaningless results. To demonstrate that the training of MoD does not suffer from fluctuations such like loss spikes, we hereby display the training losses as in Figure~\ref{fig:loss}. From the loss curves, we conclude that both MoD and MoDication are rigorously optimized.

\begin{figure}[t]
    \centering
    \includegraphics[width=0.47\textwidth]{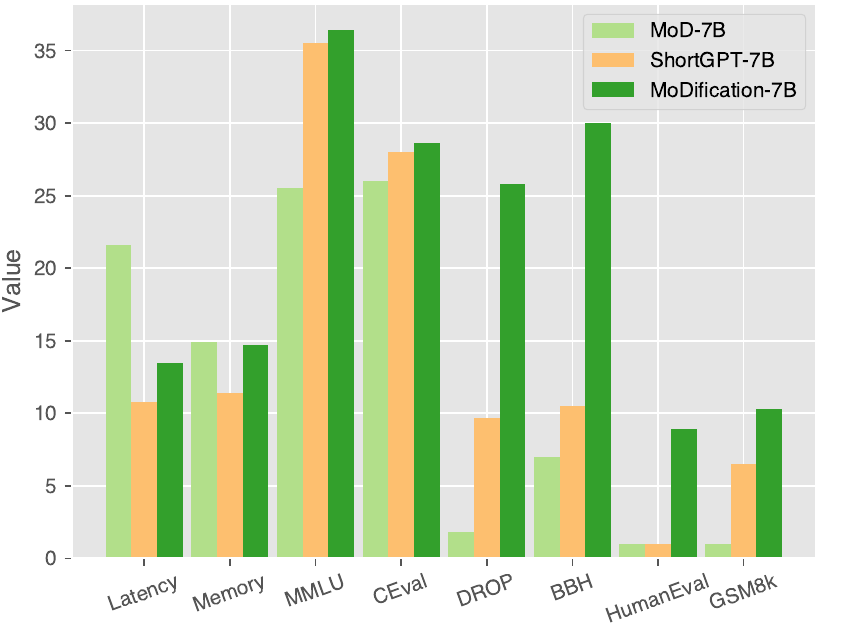}
    \caption{The comparison results concerning other efficient designs. ShortGPT is more efficient yet less effective.}
    \label{fig:shortgpt}
\end{figure}

To further enhance the plausibility of the comparison, we also compare MoDification to an alternative efficient method named ShortGPT~\citep{DBLP:journals/corr/MenXZ24} that is designed to directly drop unimportant layers, therefore decreasing latency and memory, as in Figure~\ref{fig:shortgpt}. The major conceptual difference between MoDification and ShortGPT lies in that the layer skipping of ShortGPT happens at task level. We observe that ShortGPT may lead to more apparent efficiency gains, nevertheless, it results in unacceptable performance degradation. This phenomenon implies that MoDification has considerable advantages over other efficient designs. 

\paragraph{Ablation Studies}

To get an in-depth understanding of the inner working mechanisms of MoDification, we perform a series of ablation studies on the threshold value $p$, layer loading reducing coefficient $\alpha$, etc.

\begin{figure}[ht]
    \centering
    \includegraphics[width=0.47\textwidth]{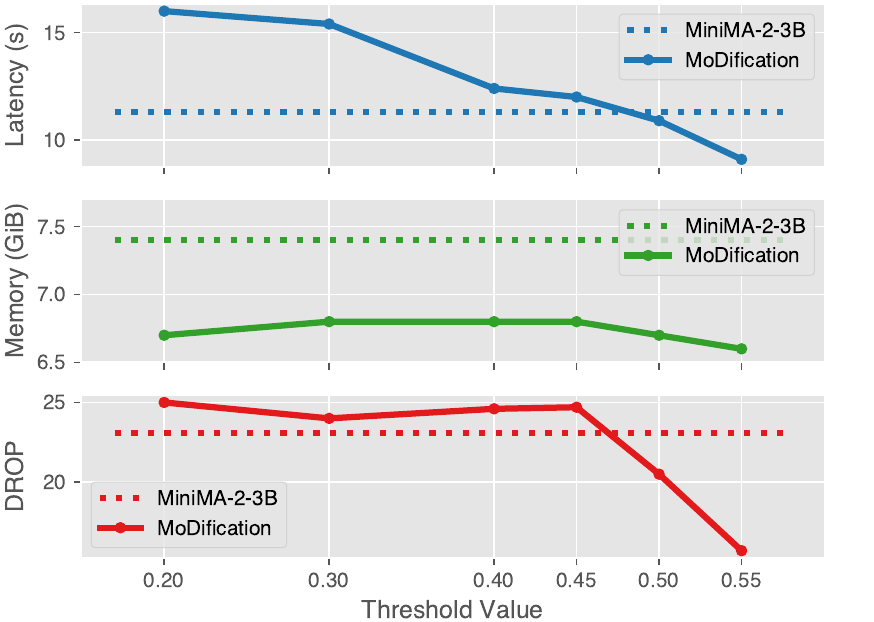}
    \caption{The impact of threshold value $p$.}
    \label{fig:threshold}
\end{figure}

\noindent \underline{\textsc{Threshold Value $p$.}} We show the impact of the threshold value $p$ in Figure~\ref{fig:threshold}, from which we inspect that while the memory consumption is always reduced across all threshold values, the latency and the performance can vary a lot from one threshold to another. Disappointingly, though smaller threshold values can cut down memory consumption, they can cause increased latency as a result of limited computation saving. For MiniMA-2-3B in the plot, the most suitable $p$ value is 0.5 regarding the trade-off between efficiency and effectiveness. However, as aforementioned, larger models such as LLaMA-2-70B urge larger $p$ values like 0.55 due to potentially larger structural sparsity. 

\noindent \underline{\textsc{Layer Load Reducing Coefficient $\alpha$.}} In Table~\ref{tab:coefficient}, we know that the layer load reducing coefficient $\alpha$ impacts both efficiency and effectiveness. Particularly, the coefficient trades effectiveness for efficiency. For instance, as $\alpha$ increases itself by an order of magnitude, the latency diminishes, so is the performance. In contrast, without the layer load reducing objective (i.e., when the coefficient is 0), MoDification tends to save very few computations. In this case, the latency is even enlarged. In summary, we assert that $\alpha$ would be better set to 0.01 to reach equilibrium.

\begin{table}[ht]
    \centering
    \caption{The impact of layer load reducing coefficient $\alpha$.}
    \begin{adjustbox}{width=0.45\textwidth,center}
    \begin{tabular}{lccc}
    \toprule
        \textbf{Method} & \makecell[c]{\textbf{Latency}\\\textcolor{gray}{Second}} & \makecell[c]{\textbf{Memory}\\\textcolor{gray}{GiB}} & \makecell[c]{\textbf{GSM8k}\\\textcolor{gray}{8-shot CoT EM}} \\
    \midrule
        MiniMA-2-3B & 11.0 & 7.4 & 14.6\\
        \quad - $\alpha$ = 0 & 15.7 & 6.7 & 8.3 \\
        \quad - $\alpha$ = 0.001 & 14.4 & 6.6 & 7.0 \\
        \quad - $\alpha$ = 0.01 & 10.9 & 6.7 & 5.7 \\
        \quad - $\alpha$ = 0.1 & 8.2 & 6.4 & 1.7 \\
    \bottomrule
    \end{tabular}
    \end{adjustbox}
    \label{tab:coefficient}
\end{table}

\noindent \underline{\textsc{Interleaved v.s. Full Strategy}} In our design, we follow the de facto strategy of original MoD and use MoDification in an interleaved style where MoDification is applied at every other layer. To confirm the interleaved strategy is the best choice, we compare the interleave startegy to a tactic that applies MoDification to top half layers and another that applies MoDification to full layers, as in Table~\ref{tab:interleave}. We note that the interleaved strategy is better than the half strategy in efficiency and is better than the full strategy in effectiveness. The reasons sit behind may be that the half strategy is inclined to conservatively skip too few computations and the full strategy is likely to skip overly many computations and affect the performance. To sum up, the interleaved strategy is currently the recommended option. 

\begin{table}[ht]
    \centering
    \caption{The impact of interleaved strategy.}
    \begin{adjustbox}{width=0.4\textwidth,center}
    \begin{tabular}{lccc}
    \toprule
        \textbf{Method} & \makecell[c]{\textbf{Latency}\\\textcolor{gray}{Second}} & \makecell[c]{\textbf{Memory}\\\textcolor{gray}{GiB}} & \makecell[c]{\textbf{MMLU}\\\textcolor{gray}{5-shot Acc}} \\
    \midrule
        Interleave & 10.9 & 6.7 & 30.0 \\
        Half & 12.8 & 6.7 & 37.4 \\
        Full & 10.5 & 6.5 & 27.4 \\
    \bottomrule
    \end{tabular}
    \end{adjustbox}
    \label{tab:interleave}
\end{table}

\noindent \underline{\textsc{Shared v.s. Separate Gate.}} We also include a refinement in MoDification where a shared gate value is multiplied to both the Attention and MLP modules if they are not skipped. Hereby, we would like to test whether two separate gate values respectively for the Attention and MLP modules are better or not. We detect that the performance would not be boosted by the separate gates but the latency would be elevated by the separate gates since additional gating is introduced. So we believe the shared gate is a fair pick. 

\begin{table}[ht]
    \centering
    \caption{The impact of shared gate.}
    \begin{adjustbox}{width=0.4\textwidth,center}
    \begin{tabular}{lccc}
    \toprule
        \textbf{Method} & \makecell[c]{\textbf{Latency}\\\textcolor{gray}{Second}} & \makecell[c]{\textbf{Memory}\\\textcolor{gray}{GiB}} & \makecell[c]{\textbf{BBH}\\\textcolor{gray}{3-shot Acc}} \\
    \midrule
        Shared & 10.9 & 6.7 & 30.8 \\
        Separate & 12.0 & 6.7 & 30.6 \\
    \bottomrule
    \end{tabular}
    \end{adjustbox}
    \label{tab:gate}
\end{table}

\paragraph{Execution Visualization}

\begin{figure}[ht]
    \centering
    \includegraphics[width=0.47\textwidth]{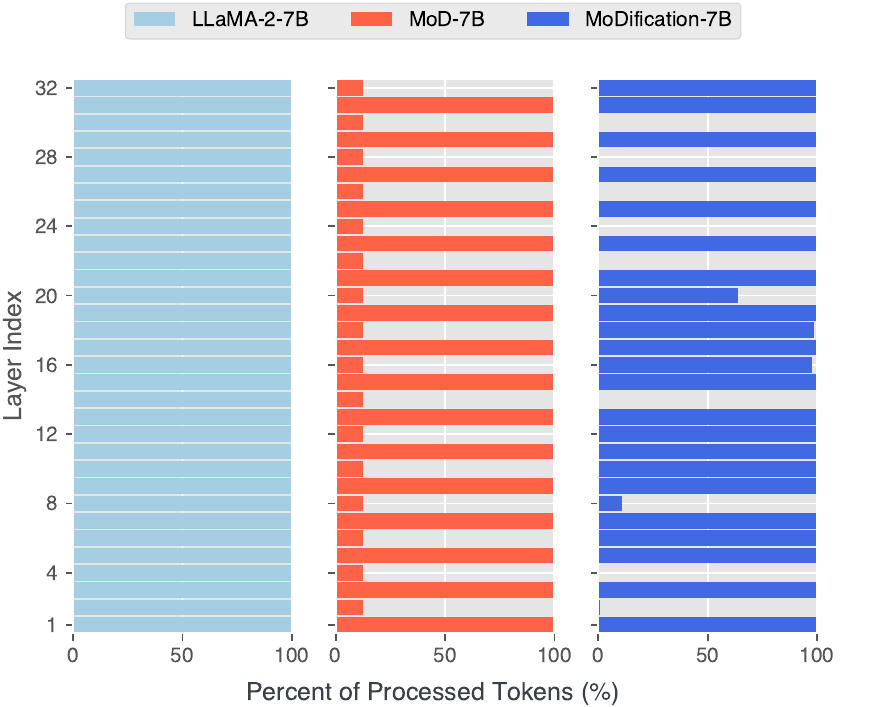}
    \caption{The visualization of executed computations of layers over tokens. MoD can save computations to a certain degree. MoDification can save computations in a more flexible way.}
    \label{fig:execution}
\end{figure}

For a more intuitive comparison between MoD and MoDification, we attempt to clearly illustrate why MoDification is more promising than MoD via visualizations of executed layers over input tokens correspondingly for MoD and MoDification.

In Figure~\ref{fig:execution}, we unleash that MoD saves computations with the sub-optimality due to the use of \textsf{top-k} operator. The forced constraint that a constant number of tokens should be retained may: 1) limit the computation saving (efficiency) when computations of more tokens should be skipped, 2) affect the computation preserving (effectiveness) when computations of fewer tokens should be skipped. Conversely, MoDification permits more flexible computation savings, leading to a better balance between efficiency and effectiveness.

\paragraph{Latency Profiling}

We previously only provide latency results in average values, and here we would like to share more latency results across all prefilling lengths and accordingly decoding lengths to offer detailed insights on how MoDification overwhelms MoD.

\begin{figure*}[ht]
    \centering
    \includegraphics[width=0.92\textwidth]{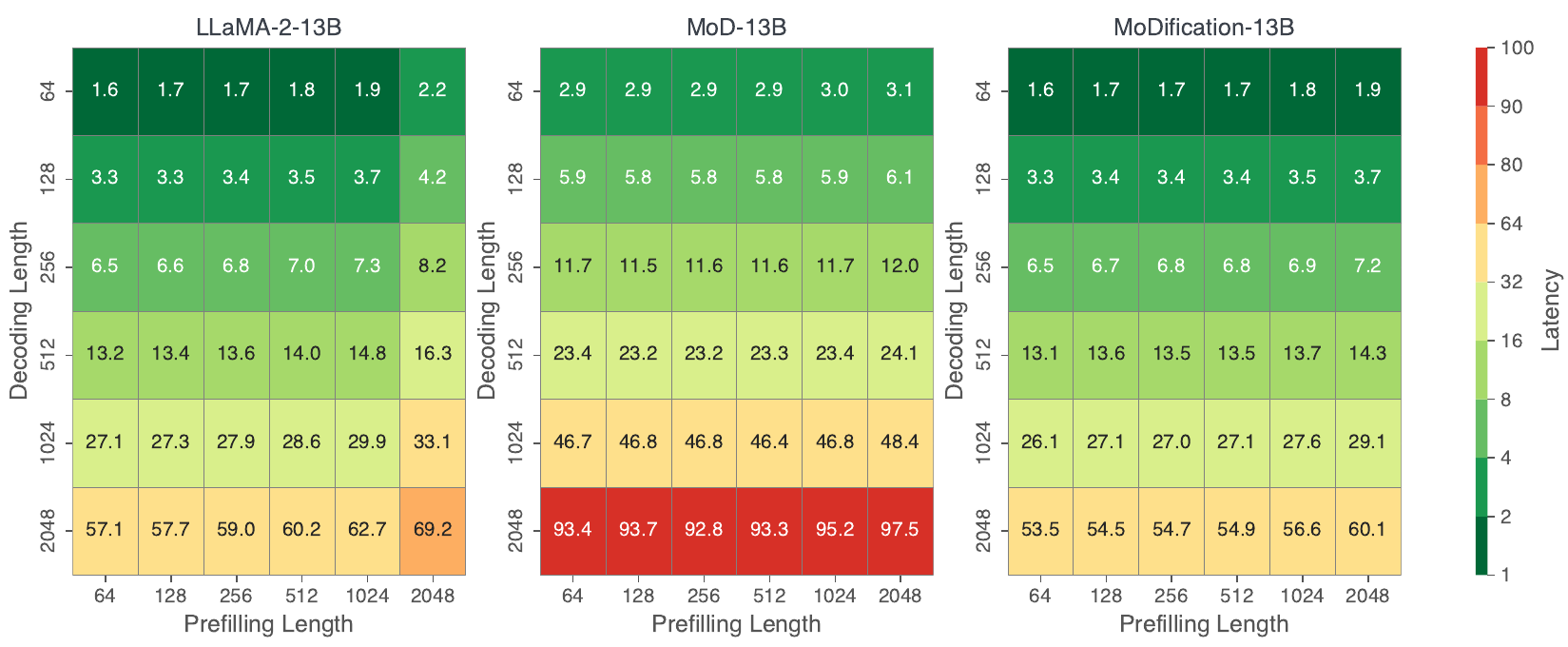}
    \caption{The detailed latency results across different prefilling lengths and decoding lengths. MoDification overwhelms both MoD and LLaMA-2 at any prefilling or decoding length.}
    \label{fig:latency}
\end{figure*}

We should tell from Figure~\ref{fig:latency} that MoDification surpasses MoD at arbitrarily any prefilling length or decoding length. And the micro gains add up to a macro gain in average. Besides, MoDification regularly overtakes LLaMA-2 in efficiency. We also find the latency is more sensitive to the decoding length increment than the prefilling length increment. Explicitly, the latency approximately doubles when the decoding length doubles but only increases a bit when the prefilling length doubles. We conjecture the distinction is resulted from that prefilling is parallel while decoding is sequential. And this hints that future explorations should be emphasized on decoding-time efficiency. Trickily, we also uncover that the latency delta along the increment of prefilling length gradually becomes larger. This pinpoints that prefilling-time efficiency would possibly be a concern when the prefilling length is immense, say more than 10k. This naturally drives us to explore the potential of MoDification in long-context applications later in Section~\ref{sec:long_ctx}.

\begin{figure}[t]
    \centering
    \includegraphics[width=0.47\textwidth]{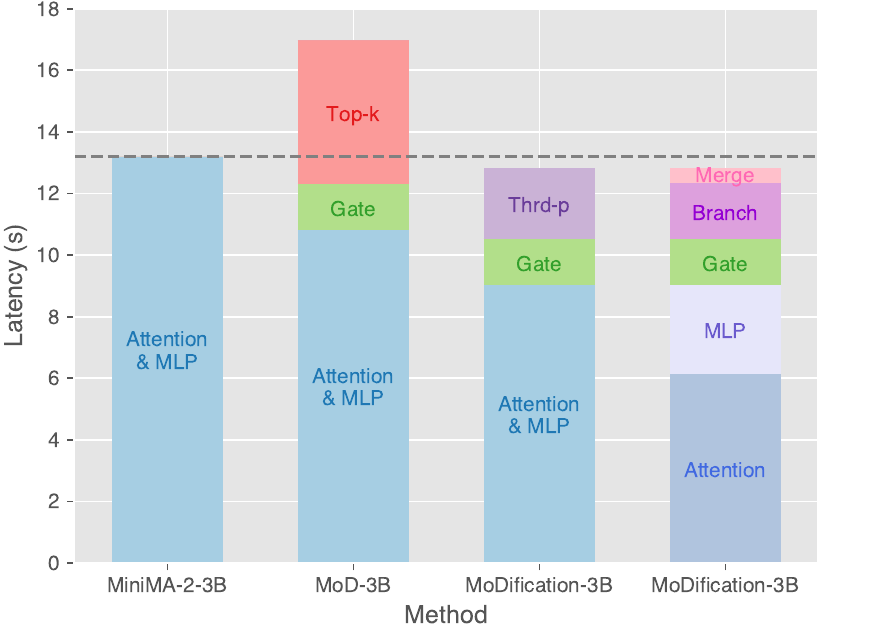}
    \caption{The nano view towards the latency.}
    \label{fig:latency_nano}
\end{figure}

More precisely, we investigate the latency through a nano view. To put it differently, we decompose the latency to three parts: the one consumed by Attention and MLP, by Gate, and by Top-k or Threshold-p. Furthermore, the Threshold-p part could be broken down to: Branch phase where some tokens are selected and the others are skipped, Merge phase where both selected and skipped tokens are merged after. The resultant chart is positioned in Figure~\ref{fig:latency_nano}.

It is manifested that MoDification not only yields smaller latency from the Thrd-p, but also yields smaller latency from the Attention and MLP. This denotes that MoDification benefits from both the transition from the Top-k to the Thrd-p and the removal of hard constraint that at least k tokens should be retained. 

\section{Long-context Applications}
\label{sec:long_ctx}

While we have fully justified the usefulness of MoDification in comparably short-context scenarios, we are obliged to substantiate the potential of MoDification in long-context applications. We take two long-context applications as testbeds, i.e., long-text generation that requires long decoding and long-context retrieval that requires long prefilling. 
As a prerequisite, we extend the context capability of MoDification-7B from 4k to 32k. We adopt NTK-aware interpolation~\citep{rozière2023code} and conduct post training on PG19 dataset~\citep{rae2019compressive}.

In long-text generation scenario where prompts sampled from Proof-Pile~\citep{Azerbayev23}, MoDification-7B can achieve up to $\sim$1.2$\times$ speedup in latency and $\sim$1.8$\times$ reduction in memory in comparison to LLaMA-2-7B. And through long-context extension, MoDification is compatible with long-context modeling and consistently behaves in low perplexity without compromise on efficiency. 

\begin{figure}[ht]
    \centering
    \includegraphics[width=0.47\textwidth]{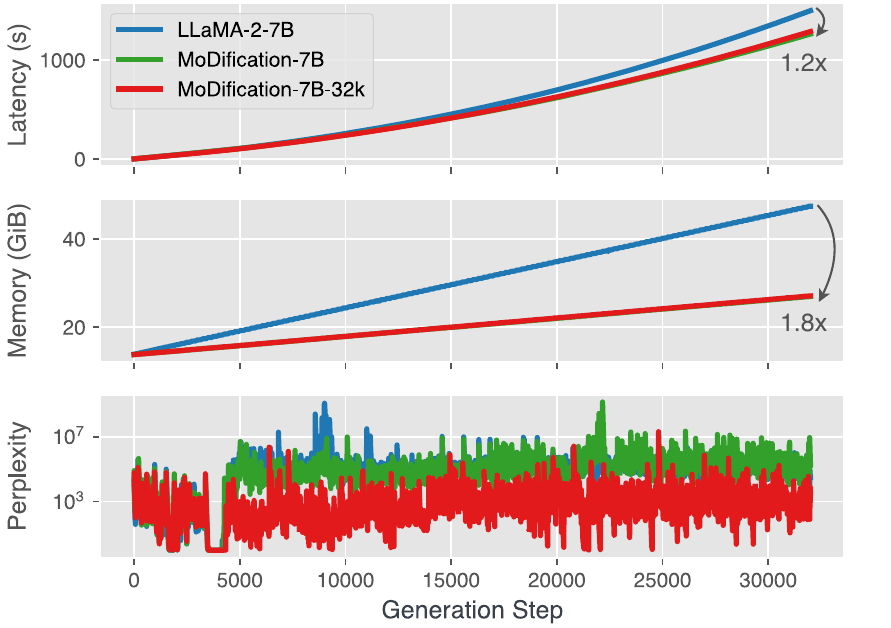}
    \caption{The application to long-text generation. MoDification-7B can achieve up to $\sim$1.2$\times$ speedup in latency and $\sim$1.8$\times$ reduction in memory in in comparison to LLaMA-2-7B. And MoDification is compatible with long-context modeling.}
    \label{fig:long_ppl}
\end{figure}

In long-context retrieval scenario where questions are gathered from Qasper~\citep{dasigi2021dataset}, MoDification-7B-32k yields 10.3 on Qasper, which is competitive with 10.2 from LLaMA-2-7B-32k. 



\section{Related Work}

\paragraph{Efficient Language Models}

Improving efficiency of language models is a long-standing and challenging task~\citep{DBLP:journals/tmlr/Wan0LA0LQYZZC024}. Besides system-level efficiency optimizations, algorithm-level efficiency optimizations mainly lie in model compression and efficient architecture. Model compression approaches~\citep{DBLP:conf/kdd/BucilaCN06} aim to reduce the model size directly. These compression approaches can be grouped into quantization~\citep{DBLP:journals/corr/DettmersLB22}, pruning~\citep{DBLP:conf/nips/MaFW23}, and distillation~\citep{DBLP:conf/emnlp/YangZS22,DBLP:conf/eacl/ZhangYWLWWS24,DBLP:conf/coling/ZhangYLW024,DBLP:conf/acl/ZhangYLWXWS23}. Efficient architectures~\citep{DBLP:journals/corr/ChildGR19,DBLP:journals/corr/Shazeer19} instead aim to reduce the quadratic time complexity of attention to quasi-linear one. Mixture of depths~\citep{DBLP:journals/corr/RaposeRR24} stands at the intersection of model compression and efficient architecture. On one hand, mixture of depths can skip layers, which can be viewed as a type of pruning. On the other hand, mixture of depths can drop tokens, which can be viewed as a sort of complexity reduction in time.

\paragraph{Conditional Computation}

A third perspective towards improving efficiency of language models would be conditional computation~\citep{DBLP:journals/corr/BengioBPP15}. It aims to relieve unnecessary computations by conditionals. This dynamic property can to the maximum extend preserve effectiveness while improving efficiency. Typical work in this area would be mixture of experts~\citep{DBLP:conf/iclr/ShazeerMMDLHD17,DBLP:journals/jmlr/FedusZS22}, early exiting~\citep{DBLP:conf/icml/ChenPLDZ24}, and mixture of depths. And mixture of depths is recently viewed as one of the best choices among available conditional computation approaches.

\section{Conclusions}

It is no doubt that MoD is an ideal choice to lift the efficiency of LLMs. In this paper, we would like to address the concern that MoD comes with a must that costly training (from scratch) should be executed. Via both architecture and data designs, we make it possible to convert existing LLMs to MoD ones with appropriate data. The designed MoDification successfully outweighs MoD in both efficiency and effectiveness.

\section*{Limitations}

Our work is limited in the following two aspects: 1) we do not apply MoDification to recently released LLMs such like LLaMA-3~\citep{DBLP:journals/corr/DubeyJP24} or Qwen-2~\citep{DBLP:journals/corr/YangYH24}, and 2) we do not apply MoDification to extremely lengthy texts, say more than 100k~\citep{DBLP:conf/acl/ZhangCHXCH0TW0024}.



\bibliography{custom}




\end{document}